\documentclass[preprint,12pt]{elsarticle}

\usepackage{amssymb}
\usepackage{amsmath}
\usepackage{amsmath,amsfonts}
\usepackage{algorithmic}
\usepackage{algorithm}
\usepackage{array}
\usepackage[caption=false,font=normalsize,labelfont=sf,textfont=sf]{subfig}
\usepackage{textcomp}
\usepackage{stfloats}
\usepackage{url}
\usepackage{verbatim}
\usepackage{graphicx}
\usepackage[table]{xcolor}
\usepackage{makecell}
\usepackage{booktabs}
\usepackage{amsthm}
\usepackage{bbm}
\usepackage{enumitem}

\journal{Pattern Recognition}
\newcommand{\zxl}[1]{\textcolor{black}{#1}}
\begin{document}

\begin{frontmatter}



\title{Implicit Counterfactual Data Augmentation \\for Robust Learning}




\author[1]{Xiaoling Zhou}
\ead{xiaolingzhou@stu.pku.edu.cn}
\author[2]{Ou Wu\corref{mycorrespondingauthor}}
\ead{wuou@ucas.ac.cn}
\author[3]{Michael K. Ng}
\ead{michael-ng@hkbu.edu.hk}
\cortext[mycorrespondingauthor]{Corresponding author.}
\address[1]{National Engineering Research Center for Software Engineering, Peking University, Beijing, 100871, China}
\address[2]{Hangzhou Institute for Advanced Study, University of Chinese Academy of Sciences, Hangzhou, 310024, China}
\address[3]{Department of Mathematics, Hong Kong Baptist University, Hong Kong, 999077, China}

\begin{abstract}
{Machine learning} models are prone to capturing the spurious correlations between non-causal attributes and classes, with counterfactual data augmentation being a promising direction for breaking these spurious associations. {However, generating counterfactual data explicitly poses a challenge, and incorporating augmented data into the training process decreases training efficiency.} This study proposes an \textbf{I}mplicit \textbf{C}ounterfactual \textbf{D}ata \textbf{A}ugmentation (ICDA) method to remove spurious correlations and make stable predictions. Specifically, first, a novel sample-wise augmentation strategy is developed that generates semantically and counterfactually meaningful deep features with distinct augmentation strength for each sample. Second, we derive an easy-to-compute surrogate loss on the augmented feature set when the number of augmented samples becomes infinite. Third, two concrete schemes are proposed, including direct quantification and meta-learning, to derive the key parameters for the robust loss. In addition, ICDA is explained from a regularization perspective, {revealing its capacity to improve intra-class compactness and augment margins at both class and sample levels.} Extensive experiments have been conducted across various biased learning scenarios covering both image and text datasets, demonstrating that ICDA consistently enhances the generalization and robustness performance of popular networks.
\end{abstract}




\end{frontmatter}



\section{Introduction}
\begin{figure}[t] 
\centering
\includegraphics[width=0.85\textwidth]{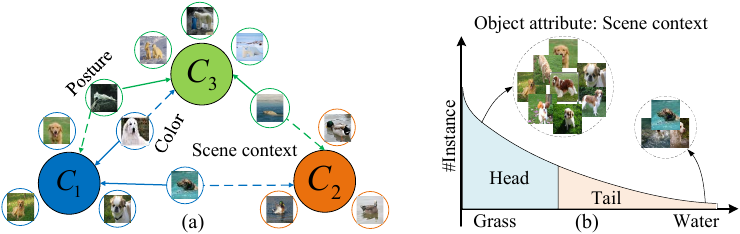}
\caption{(a): Illustration for images affected by spurious correlations due to rare attributes (e.g., rare posture, rare color, and rare scene context). $C_1$, $C_2$, and $C_3$ are the dog, drake, and polar bear classes, respectively. The solid line connects the sample's ground-truth class, and the dotted line connects the class with a spurious correlation with the sample. (b): Illustration for attribute imbalance. \zxl{Regarding the attribute of scene context, the majority of dogs in the training data are situated on grass, while only a small number are depicted in water. 
Imbalances in attributes generally lead to spurious correlations between non-causal attributes and labels in deep learning models.}}
\label{fig1}
\end{figure}

{D}{eep} learning models are supposed to learn invariances and make stable predictions based on some right causes. However, models trained with empirical risk minimization are prone to learning spurious correlations and suffer from high generalization errors when the training and test distributions do not match~\cite{r2,zhang2025robust}. For example, dogs are mostly on the grass in the training set. Thus, a dog in the water can easily be misclassified as a ``drake" due to its rare scene context (``water") in the ``dog" class, \zxl{which is }illustrated in Fig.~\ref{fig1}. A promising solution for enhancing the generalization and robustness of deep learning models is to learn causal representations~\cite{r45}. If a model can focus more on causal correlations rather than spurious associations, it is more likely to produce stable and accurate predictions. 

Counterfactual augmentation has become popular for causal models because of \zxl{its capacity to enhance model robustness} and being model-agnostic. For instance, Cao et al.~\cite{cao2025enhancing} combined self-supervised and contrastive learning for unbiased training.
Moreover, Chang et al.~\cite{r1} introduced two new image generation procedures that included counterfactual and factual data 
augmentations to reduce spuriousness between backgrounds of images and labels, achieving higher accuracy in several challenging datasets. Mao et al.~\cite{r2} utilized a novel strategy to learn robust representations that steered generative models to manufacture interventions on features caused by confounding factors. Nevertheless, the methods presented above suffer from several shortcomings. Specifically, it is not trivial to explicitly distinguish between causal and non-causal attributes, and the training efficiency will decline as excess augmented images are involved in training. 

It should be mentioned that implicit data augmentation settles the inefficiency of explicit augmentation by avoiding the generation of excess samples.
Implicit Semantic Data Augmentation (ISDA)~\cite{r8} conducts a pioneering study on implicit data augmentation. \zxl{It is inspired by the observation that deep features in a network are typically linearized, resulting in the existence of numerous semantic directions in the deep feature space.} Then, it translates samples along the semantic directions in the feature space based on an assumed class-wise augmentation distribution. By deriving an upper bound on the expected cross-entropy (CE) loss,
ISDA enables optimization of only the upper bound to achieve data augmentation in an efficient way. \zxl{Subsequent studies on imbalance learning have expanded upon this approach. For instance, MetaSAug~\cite{r9} optimizes the covariance matrix of the tail classes on a balanced metadata set to mitigate the issue of inaccurate estimation arising from the insufficient number of samples in the tail classes,} 
yielding good performance on imbalanced data. Besides, \zxl{to generate more diverse samples for tail classes,}
Reasoning-based Implicit Semantic Data Augmentation (RISDA)~\cite{r79} \zxl{augments samples in tail classes using semantic vectors from not only the current class but also the relevant classes. However, these methods, specifically designed for imbalanced learning, may not effectively dismantle the spurious associations within deep learning models. Moreover, }they adopt purely class-wise semantic augmentation strategies, and thus samples in the same class have identical augmentation distributions that are inaccurate and non-specific. 
\zxl{As illustrated in} Fig.~\ref{fig1}(a), samples in the same class \zxl{may exhibit spurious correlations with different classes due to various attributes. Consequently, an ideal augmentation strategy should consider these sample-wise non-causal attributes.}

This study proposes a \zxl{novel} sample-wise \textbf{I}mplicit \textbf{C}ounterfactual \textbf{D}ata \textbf{A}ugmentation (ICDA) method that facilitates both semantic and counterfactual augmentations. Semantic augmentation is accomplished by transforming samples along vectors drawn from the deep feature space of the ground-truth class. Moreover, counterfactual augmentation is realized by manipulating samples along vectors sourced from the deep feature spaces of non-target classes.  
The augmentation distribution and strength for each sample are determined based on class-wise statistical information and the degree of spurious correlation between the sample and each class. \zxl{Then, we verify that ICDA approximates a novel robust surrogate loss (termed the ICDA loss) by considering the number of augmentations becoming infinite, making the process highly efficient.} Furthermore, meta-learning is introduced to learn key parameters in this novel loss, which is analyzed and compared against existing methods in a unified regularization perspective, revealing that it enforces extra intra-class compactness by reducing the classes' mapped variances and encourages larger sample 
margins and class-boundary distances. Extensive experiments verify that ICDA consistently achieves state-of-the-art performance in several typical learning scenarios requiring the models to be robust and presenting a high generalization ability. Furthermore, the visualization results indicate that ICDA generates more diverse and meaningful counterfactual images with rare attributes, helping models break spurious correlations and affording stable predictions for the right reasons.

\section{Related Work}


\subsection{Data Augmentation}
Data augmentation approaches are popular for enhancing the generalization and robustness of deep learning models~\cite{r8,li2023towards}. 
Counterfactual augmentation generates hypothetical samples (i.e., counterfactuals) by making small changes to the original samples, which can be divided into hand-crafted~\cite{r1,r9} 
and using causal generative models~\cite{r55}, demonstrating competitive performance. However,
existing counterfactual data augmentation approaches predominantly rely on explicit augmentation methods, while explicitly identifying non-causal attributes presents significant challenges, and training models on such augmented data often leads to inefficiencies~\cite{r55}. Implicit semantic data augmentation~\cite{r8,r9,r79} overcomes the inefficiency of explicit data 
augmentation approaches as it does not generate excess samples and achieves the effect of augmentation only through optimizing the surrogate loss when the number of augmented samples becomes infinite, which is thus efficient. However, implicit data augmentation can not help overcome spurious associations.
Besides, all samples or samples in the same class employ the same augmentation strategy, including both augmentation distribution and strength, which is naturally not optimal. 

\subsection{Logit Adjustment}
Logit vectors represent the outputs before the Softmax layer in the majority of deep classifiers. Logit adjustment approaches involve introducing perturbation terms to the logit vector, aiming to bolster the robustness of deep learning models. This technical path was originally proposed in face recognition~\cite{r28}, seeking to increase inter-class distance and intra-class compactness. Presently, logit adjustment is employed, implicitly or explicitly, in various contexts such as data augmentation~\cite{r8,r79} and long-tailed classifications~\cite{r30,r32}. \zxl{For example, the Logit-adjusted (LA)~\cite{r32} loss incorporates class proportion terms as perturbations, demonstrating effectiveness in imbalanced learning scenarios. Additionally, ISDA~\cite{r8} can be considered a logit adjustment approach, contributing significantly to the enhancement of models' generalization capabilities.}

\begin{figure}[t] 
\centering
\includegraphics[width=0.95\textwidth]{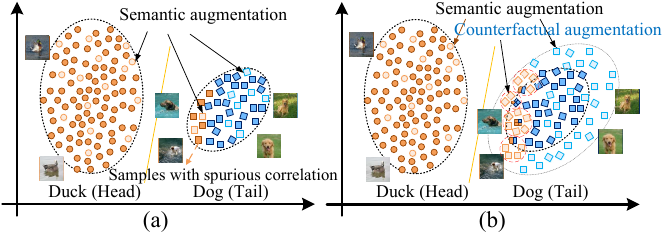}
\vspace{-0.1in}
\caption{(a): Diagram for ISDA, which only conducts semantic augmentation and treats all samples equally. (b): Diagram for ICDA, containing  
both semantic and counterfactual augmentations. Samples in the tail class (i.e., Dog class) and those with rare attributes (i.e., dogs in the water) are augmented the most. 
The two axes mean
the dimensions of the 2D feature space, in which each sample is represented by a dot or square.
Solid and transparent samples are the original and augmented ones. Samples in the same red circle are augmented from the same sample. The augmentation strength is determined by the degree of spurious associations.}
\label{fig3}
\end{figure}

\section{Implicit Counterfactual Data Augmentation}
\textbf{Notation.} Consider training a network $G$ with weights
$\boldsymbol{W}$ on a training set $D^{{train}}=\{(\boldsymbol{x}_{i},y_{i})\}_{i=1}^{N}$, where $y_{i}\in\{1,\cdots,{C}\}$ is the label of the $i$th sample $\boldsymbol{x}_{i}$ over ${C}$ classes. Let the $H$-dimensional vector $\boldsymbol{h}_{i}=G(\boldsymbol{x}_{i},\boldsymbol{W})$ denote the deep feature of $\boldsymbol{x}_{i}$
learned by $G$. Let $\boldsymbol{u}_{i}=f(\boldsymbol{h}_{i}) =\boldsymbol{w}\boldsymbol{h}_{i}+\boldsymbol{b}$ denote the logit vector, $\boldsymbol{w}= [\boldsymbol{w}_{1}, \cdots, \boldsymbol{w}_{{C}}]^{\zxl{T}} \in\mathrm{R}^{C\times H}$, and $\boldsymbol{b}= [{b}_{1}, \cdots, {b}_{{C}}]^{\zxl{T}}\in\mathrm{R}^{C}$. Let $\boldsymbol{\mu}_c$ and $\boldsymbol{\Sigma}_c$ be the mean and covariance matrix of the \zxl{deep} features for class $c$. $\mathcal{N}(\boldsymbol{\mu}, \boldsymbol{\Sigma})$ means a multivariate normal distribution with mean vector $\boldsymbol{\mu}$ and covariance matrix $\boldsymbol{\Sigma}$.

\subsection{\zxl{Counterfactual Data Augmentation}}
\zxl{To mitigate spurious correlations between non-causal attributes and classes, we propose a counterfactual data augmentation strategy, which generates both meaningful semantic and counterfactual samples. 
Considering that the spurious correlations between samples and classes are sample-wise, we devise and utilize sample-level augmentation distributions. To achieve semantic augmentation, perturbation vectors for the deep feature of each sample, $\boldsymbol{h}_{i}$, are sampled from a multivariant normal distribution $\mathcal{N}(\boldsymbol{0},\boldsymbol{\Sigma}_{y_i})$}. 
To mitigate spurious correlations, we intervene on non-causal attributes that are spuriously correlated with other classes, while preserving the core object features to generate counterfactual instances.
Specifically, the deep features of samples $\boldsymbol{h}_{i}$ are transformed along the perturbation vectors extracted from the deep feature spaces of non-target classes, i.e., $\mathcal{N}(\boldsymbol{\mu}_{c},\boldsymbol{\Sigma_{{c}}})$, where $c\neq y_i$.
\zxl{
Consequently, when augmenting the deep feature $\boldsymbol{h}_{i}$ to class $c$, the perturbation vectors are sampled from $\mathcal{N}(\boldsymbol{0}+\alpha_{i,c}\boldsymbol{\mu}_{c},\boldsymbol{\Sigma}_{y_{i}}+\alpha_{i,c}\boldsymbol{\Sigma_{{c}}})$, where $\alpha_{i,c}$ $(\geq 0)$ is determined by the degree of the spurious association between $\boldsymbol{x}_{i}$ and class $c$.}

\zxl{As for the augmentation strength, that is the number of augmented samples $\tilde{M}_{i,c}$ for sample $\boldsymbol{x}_{i}$ to class $c$, it is assumed to follow $\tilde{M}_{i,c}=(M{\alpha_{i,c}})/{\pi_{y_{i}}}$, where $\pi_{y_{i}}$ denotes the proportion of class $y_i$ and $M$ is a constant. Consequently, the higher the degree of the spuriousness between $\boldsymbol{x}_{i}$ and class $c$ and the smaller the $\pi_{y_{i}}$, the larger the number ($\tilde{M}_{i,c}$) of samples will be augmented from $\boldsymbol{x}_{i}$ to class $c$.}  \zxl{Fig.~\ref{fig3}(a) highlights that the existing augmentation approaches ignore the relationship between samples and the other class, and all samples are treated equally, prohibiting a well-adjusted distribution.}
Fig.~\ref{fig3}(b) \zxl{presents our augmentation manner, in which} samples in the tail class and the ones most spuriously correlated with attributes of the other class are augmented most, \zxl{facilitating enhancing the generalization and robustness of models against spurious correlations.}

During training, $C$ feature means and covariance matrices are computed, one for each class. To enhance efficiency, the values of $\boldsymbol{\mu}_{c}$ and $\boldsymbol{\Sigma}_{c}$ are computed online by aggregating statistics from all mini-batches, which are given in Appendix~A. Given that the estimated statistics information in the first few epochs is not quite informative, we add a scale parameter $\lambda=(t/\mathcal{T})\times \lambda^{0}$ before the estimated $\boldsymbol{\mu}$ and $\boldsymbol{\Sigma}$, where $t$ and $\mathcal{T}$ refer to the numbers of the current and total iterations. Additionally, $\lambda_{0}$ is a hyperparameter. The augmented feature ${\boldsymbol{h}}_{i,c}$ for $\boldsymbol{h}_{i}$ to class $c$
is obtained by translating $\boldsymbol{h}_{i}$ along a random direction sampled from the above multivariate normal distribution. Equivalently, we have $\boldsymbol{{h}}_{i,c}\sim\mathcal{N}(\boldsymbol{{h}}_{i} +\lambda\alpha_{i,c}\boldsymbol{\mu}_{c},\lambda(\boldsymbol{\Sigma}_{y_{i}}+\alpha_{i,c}\boldsymbol{\Sigma}_{{c}}))$.

Notably, \zxl{our augmentation strategy} has distinct differences from current \zxl{semantic} augmentation methods:
\begin{itemize}[itemsep=2pt, parsep=0pt]
\item 
Their motivations are different. \zxl{Our strategy} aims to generate more \zxl{counterfactual} data for 
breaking spurious correlations, while the existing methods \zxl{only} generate diverse semantic data.
\item Their granularities are different. \zxl{Our} augmentation strategy is sample-wise, which is fine-grained and pinpoint, while current schemes involve class-wise strategies. 
\item \zxl{Our strategy} highlights the augmentation strength, which is crucial in an augmentation strategy, as inappropriate class and attribute distributions always cause spuriousness. However, it is overlooked by the existing methods.
\end{itemize}

\subsection{New Robust Loss under Implicit Augmentation}
\label{sec32}
\zxl{A naive method to implement ICDA is to explicitly augment the deep features of samples based on the designed augmentation distribution and strength.} Specifically, for class $c$ ($\neq y_i$), the deep features $\boldsymbol{h}_{i}$ are augmented $\tilde{M}_{i,c}$ times utilizing perturbation vectors sampled from the corresponding distribution $\mathcal{N}(\boldsymbol{0}+\lambda\alpha_{i,c}\boldsymbol{\mu}_{c},\lambda(\boldsymbol{\Sigma}_{y_{i}}+\alpha_{i,c}\boldsymbol{\Sigma_{{c}}}))$. Consequently, an
augmented feature set $\{\{\boldsymbol{h}_{i,c}^{1},\cdots, \boldsymbol{h}_{i,c}^{\tilde{M}_{i,c}}\}_{c=1, c\neq y_{i}}^{{C}}\}_{i=1}^{N}$ can be formed. 
Then, the corresponding CE loss for all augmented features is
\begin{equation}
\begin{aligned}
\mathcal{L}_M(\boldsymbol{w}, \boldsymbol{b}, \boldsymbol{W})=\frac{1}{\tilde{M}}\sum_{i=1}^{N}\sum_{c\neq y_{i}}\sum_{k=1}^{\tilde{M}_{i,c}}-\log\frac{\exp[{f_{y_{i}}(\boldsymbol{h}_{i,c}^{k})}]}{\sum_{j=1}^{{C}}\exp[{f_{j}(\boldsymbol{h}_{i,c}^{k})]}},
\end{aligned}
\end{equation}
where $\tilde{M}\!=\! \sum_{i=1}^{N}\!\sum_{c=1,c\neq y_{i}}^{{C}}\!\tilde{M}_{i,c}$ and $f_{j}(\boldsymbol{h}_{i,c}^{k})=\boldsymbol{w}_{j}^{T}\boldsymbol{h}_{i,c}^{k}+{b}_{j}$. \zxl{To augment more data while enhancing training efficiency,} we let $M$ in $\tilde{M}_{i,c}$ grow to infinity. Then, the expected CE loss for all augmented features is
\begin{equation}
\begin{aligned}
\mathcal{L}_\infty(\boldsymbol{w}, \boldsymbol{b}, \boldsymbol{W})
    &=\frac{1}{\tilde{N}}\sum_{i=1}^{N}\sum_{c\neq y_{i}}\tilde{N}_{i,c}\mathbbm{E}_{\boldsymbol{h}_{i,c}}[-
    \log\frac{\exp({f_{y_{i}}(\boldsymbol{h}_{i,c})})}{\sum_{j=1}^{{C}}\exp({f_{j}(\boldsymbol{h}_{i,c})})}],
    \end{aligned}
    \label{eq2}
\end{equation}
where $\tilde{N}_{i,c}\!=\!{\alpha_{i,c}}/{\pi_{y_{i}}}$ and $\tilde{N}\!=\!\sum_{i=1}^{N}\sum_{c=1,c\neq y_{i}}^{{C}}\tilde{N}_{i,c}$. \zxl{However, the above expected CE loss is hard to calculate.} Then, we derive a more easy-to-compute surrogate loss for Eq.~(\ref{eq2}), which is as follows:
\begin{equation}
    \begin{aligned}
    \mathcal{L}_{s}(\boldsymbol{w}, \boldsymbol{b}, \boldsymbol{W})& =
    \frac{1}{\tilde{N}}\sum_{i=1}^{N}\frac{1}{\pi_{y_{i}}}\log(1\!+\!\sum_{c\neq y_{i}}\exp[f_{c}(\boldsymbol{h}_{i})\!-\!f_{y_{i}}(\boldsymbol{h}_{i})\!+\!\phi_{i,c}]),
    \\
    \phi_{i,c} &= ({{\lambda}}/{2})P_{c,{i}}+{\lambda} Q_{c,{i}}+\beta \alpha_i, \end{aligned}
\end{equation}
where $P_{c,{i}}\!=\!\Delta \boldsymbol{w}_{c,{y_i}}(\boldsymbol{\Sigma}_{y_{i}}\!+\!\sum_{j=1, j\neq y_{i}}^{{C}}\hat{\alpha}_{i,j}\boldsymbol{\Sigma}_{{j}})\Delta \boldsymbol{w}_{{c},y_{i}}^{T}$ and $Q_{c,{i}} \!=\!\Delta \boldsymbol{w}_{c,{y_i}}\sum_{j=1,j\neq y_{i}}^{{C}}\hat{\alpha}_{i,j}\boldsymbol{\mu}_{j}$, in which $\Delta \boldsymbol{w}_{c,y_{i}}\!=\! \boldsymbol{w}_{c}^{T}\!-\!\boldsymbol{w}_{y_{i}}^{T}$ and $\hat{\alpha}_{i,j}={\alpha_{i,j}}/{({C}\!-\!1)}$. In addition, $\alpha_{i}\!=\!\sum_{j=1,j\neq y_{i}}^{{C}}{\hat{\alpha}}_{i,j}$. The inference details are presented in Appendix~B. \zxl{Consequently, instead of conducting the augmentation process explicitly, we can directly optimize this surrogate loss.}

Although $\mathcal{L}_{s}(\boldsymbol{w}, \boldsymbol{b}, \boldsymbol{W})$ can be directly utilized during training, a more effective loss is leveraged after adopting the two following modifications: (1) Inspired by the manner in LA~\cite{r32}, the class-wise weight $1/\pi_{y_i}$ is replaced by a perturbation term on logits. (2) We only retain the term 
$\Delta \boldsymbol{w}_{{c},y_{i}}\hat{\alpha}_{i,c}\boldsymbol{\mu}_{c}$ in $Q_{c,{i}}$. The reason for the proposed variation is detailed in Section~\ref{secregu}. Accordingly, the final ICDA training loss becomes
\begin{equation}
    \begin{aligned}
\overline{\mathcal{L}}_{s}(\boldsymbol{w}, \boldsymbol{b}, \boldsymbol{W})& = 
    \frac{1}{\tilde{N}}\sum_{i=1}^{N}\log(1\!+\!\sum_{c\neq y_{i}}\exp[f_{c}(\boldsymbol{h}_{i})\!-\!f_{y_{i}}(\boldsymbol{h}_{i})\!+\!\hat{\phi}_{i,c}]),
    \\ \hat{\phi}_{i,c} &= ({{\lambda}}/{2}) P_{c,{i}}+{\lambda} \Delta \boldsymbol{w}_{c, y_{i}}\hat{\alpha}_{i,c}\boldsymbol{\mu}_{c}+\delta_{c,{i}}+\beta\alpha_{i}, 
       \end{aligned}
\end{equation}
where $\delta_{c,{i}}=\log({\pi_{c}}/{\pi_{y_{i}}})$ and $\beta$ is a hyperparameter which is fixed as $0.1$ in our experiments. \zxl{Notably, the ICDA loss can be considered as a generalization of several typical logit adjustment losses.} For example, when ${\lambda}=\beta=0$, our method can be reduced to LA. Additionally, for $\hat{\alpha}_{i,c}=\beta=0$ (\zxl{$c\neq y_i$}) and balanced classes, ICDA degenerates to ISDA. 
Section~\ref{secregu} further demonstrates the superiority of the proposed ICDA loss over existing methods from the perspective of regularization.

\begin{figure*}[t] 
\centering
\includegraphics[width=1\textwidth]{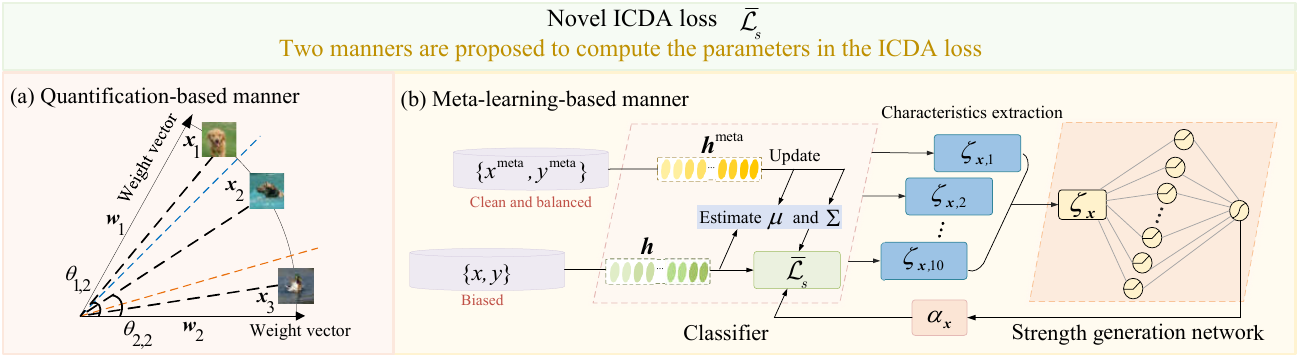}
\vspace{-0.2in}
\caption{Two manners for applying the proposed ICDA loss: quantification-based manner and meta-learning-based manner. (a): Illustration for the angle between the sample feature and the weight vector. The angle between the deep feature of $\boldsymbol{x}_{2}$ and the classifier weights for class $C_{2}$ is smaller compared to the angle between the deep feature of $\boldsymbol{x}_{1}$ and the classifier weights for class $C_{2}$, attributed to the spurious correlation between $\boldsymbol{x}_{2}$ and $C_{2}$. (b): The overall structure of Meta-ICDA, which comprises three main components: the classifier, the characteristics extraction module, and the strength generation network.}
\label{fig4}
\end{figure*}


\section{Learning with ICDA}
When applying the derived ICDA loss to optimize the classifier, it is first necessary to determine the calculation of several hyperparameters—namely, $\boldsymbol{\mu}_c$, $\boldsymbol{\Sigma}_c$, and $\alpha_{i,c}$—which govern the direction and strength of data augmentation.
\zxl{Therefore, two approaches are proposed to optimize the classifier using the ICDA loss: a direct quantification-based method and a meta-learning-based method, as illustrated in Fig.~\ref{fig4}.}

\subsection{Direct Quantification-Based Manner}
The spurious correlation between sample $\boldsymbol{x}_{i}$ and class $c$ can be directly quantified by the angle ($\theta_{i,c}$) between $\boldsymbol{h}_i$ and the weight vector $\boldsymbol{w}_c$ of class $c$. Naturally, 
the larger the spurious correlation between $\boldsymbol{h}_{i}$ and class $c$, the smaller the $\theta_{i,c}$ and the larger the $\cos\theta_{i,c}$. An illustration is presented in Fig.~\ref{fig4}(a). Samples $\boldsymbol{x}_{1}$ and $\boldsymbol{x}_{2}$ both belong to class $C_{1}$. Nevertheless, $\theta_{2,2}$ is smaller than $\theta_{1,2}$ as $\boldsymbol{x}_2$ is more spuriously correlated with class $C_2$. 

Since $\alpha_{i,c}$ is determined by the degree of spurious correlation between $\boldsymbol{x}_{i}$ and class $c$, it should be positively correlated with $\cos\theta_{i,c}$. Moreover, only when the direction of $\boldsymbol{h}_{i}$ is partially consistent with that of $\boldsymbol{w}_{c}$ (i.e., $\theta_{i,c}<90^{\circ}$), the information of class $c$ should be utilized to augment sample $\boldsymbol{x}_{i}$.
Thus, we denote $\alpha_{i,c}=\max(\cos\theta_{i,c},0)$, where a larger $\alpha_{i,c}$ value means a larger counterfactual augmentation strength.  
Then, we have $\alpha_{i}\propto\sum_{c\neq y_{i}}\max(\cos\theta_{i,c},0)$. Nevertheless, quantifying $\alpha_{i}$ through angle 
$\theta_{i,y_i}$ is more direct. If $\boldsymbol{x}_{i}$ is notably influenced by the spurious correlations with other classes, then $\theta_{i,y_{i}}$ will be large, and $\cos\theta_{i,y_i}$ will be small. Thus, $\alpha_{i}$ should be negatively correlated with $\cos\theta_{i,y_i}$. Meanwhile, the value range of $\alpha_{i}$ is 
restricted to $[0, 1]$. Therefore, we let $\alpha_{i} = {(1-\cos\theta_{i,y_{i}})}/{2}$.
This manner is empirically verified to be 
more effective. 

%


\subsection{Meta-Learning-Based Manner}

If metadata are available, the extent a sample is affected by spurious correlation can be better determined by training a strength generation network. \zxl{The manner based on meta-learning is called Meta-ICDA.} The input of the strength generation network involves ten training characteristics of samples $\boldsymbol{\zeta}_{i}$, including loss, margin, uncertainty, etc., denoted by ${\zeta}_{{i},1}, \cdots, {\zeta}_{{i},10}$. Details for the extracted training characteristics are presented in Appendix~C. In this study, the network is a two-layer MLP, and its output is the augmentation strength $\alpha_{i}$. Thus, we have $\alpha_{i} = \text{MLP}(\boldsymbol{\zeta}_{i})$.
Considering that the geodesic distance between the sample and other classes can well measure their correlation, $\alpha_{i,c}$ is still calculated by $\max(\cos\theta_{i,c},0)$. The estimated covariance matrices and the feature means are optimized on metadata because biased training 
data (e.g., imbalanced and noisy data) may not estimate the statistical information well. Fig.~\ref{fig4}(b) illustrates the framework of Meta-ICDA, which includes three main parts: the classifier network, the strength generation network, and the characteristics extraction module. We utilize an online meta-learning-based learning strategy to alternatively update the parameters of the classifier and the strength generation network. 
The optimization process is detailed below.


To ease this paper's notation, the deep classifier network's parameters of $\boldsymbol{W}$ and $\boldsymbol{w}$ are denoted as $\tilde{\boldsymbol{W}}$. The deep classifier which includes both feature extractor $G$ and classifier $f$ is denoted as $\tilde{f}$. The parameters in the strength generation network are $\boldsymbol{\Omega}$. The small metadata set is denoted as $D^{{meta}}=\{(\boldsymbol{x}_{i}^{{meta}},y_{i}^{{meta}})\}_{i=1}^{B}$, where $B\le N$. 

\begin{algorithm}[t]
    \caption{Meta-ICDA}
    \label{alg1}
    \textbf{Input}: Training data $D^{train}$, metadata ${D}^{meta}$, batch sizes $n$ and $m$, step sizes $\eta_{1}$ and $\eta_{2}$, number of iterations $\mathcal{T}$.\\
       \textbf{Output}: Learned $\tilde{\boldsymbol{W}}$ and $\boldsymbol{\Omega}$.
    \begin{algorithmic}[1]
       \STATE {Initialize $\tilde{\boldsymbol{W}}^{(1)}$ and $\boldsymbol{\Omega}^{(1)}$;}
\FOR {$t = 1$ to $\mathcal{T}$}
    \STATE{Sample 
    $ \{(\boldsymbol{x}_{i},{y}_{i})\}_{i=1}^{n}$ from $D^{train}$;}
    \STATE{Sample 
    $\{(\boldsymbol{x}^{meta}_{i},{y}^{meta}_{i})\}_{i=1}^{m}$ form $D^{meta}$;}
    \STATE{Calculate current feature means $\boldsymbol{\mu}^{(t)}$ and covariance matrices $\boldsymbol{\Sigma}^{(t)}$;}
    \STATE{Formulate 
    $\overline{\tilde{\boldsymbol{W}}}^{(t)}(\boldsymbol{\Omega})$ by Eq.~(\ref{meta1});}
    \STATE{Update 
    $\boldsymbol{\Omega}^{(t+1)}$ by Eq.~(\ref{meta2});}
    \STATE{Update $\boldsymbol{\mu}^{(t+1)}$ and $\boldsymbol{\Sigma}^{(t+1)}$ by Eqs.~(\ref{meta3}) and (\ref{meta4});}
    \STATE{ Update 
    $\tilde{\boldsymbol{W}}^{(t+1)}$ by Eq.~(\ref{meta5});}
    
    \ENDFOR
    \end{algorithmic}
\end{algorithm}

During this process, first, $\boldsymbol{\Omega}$ is treated as the to-be-updated parameter, and 
the parameter of the deep classifier $\tilde{f}$, that is $\tilde{\boldsymbol{W}}$, which is a function of $\boldsymbol{\Omega}$, 
is formulated. We utilize the stochastic gradient descent (SGD) optimizer to optimize the training loss on a minibatch of training samples $\{(\boldsymbol{x}_{i},{y}_{i})\}_{i=1}^{n}$ in each iteration, where $n$ is the size of the mini-batch. 
Thus, $\tilde{\boldsymbol{W}}$ is formulated by the following equation:
\begin{equation}
\begin{aligned}
\overline{\tilde{\boldsymbol{W}}}^{(t)}(\boldsymbol{\Omega})&=\tilde{\boldsymbol{W}}^{(t)}-\eta_{1} \frac{1}{n} \sum_{i=1}^{n} \nabla_{_{\tilde{\boldsymbol{W}}}}\ell_{ICDA}(\tilde{f}(\boldsymbol{x}_{i}),y_{i}; \alpha_{i}^{(t)})|_{_{\tilde{\boldsymbol{W}}^{(t)}}},
\end{aligned}
\label{meta1}
\end{equation}
where $\eta_{1}$ is the step size. After extracting the training characteristics from the classifier, the parameters of the strength generation network
$\boldsymbol{\Omega}$ 
can be updated on a minibatch of metadata $\{(\boldsymbol{x}_{i}^{{meta}},{y}_{i}^{{meta}})\}_{i=1}^{m}$ as follows:
\begin{equation}
\begin{aligned}
\boldsymbol{\Omega}^{(t+1)}&= \boldsymbol{\Omega}^{(t)}-\eta_{2} \frac{1}{m} \sum_{i=1}^{m} \nabla_{\boldsymbol{\Omega}}\ell_{CE}(\tilde{f}_{\overline{\tilde{\boldsymbol{W}}}(\boldsymbol{\Omega}^{(t)})}(\boldsymbol{x}_{i}^{{meta}}),y_{i}^{{meta}})|_{\boldsymbol{\Omega}^{(t)}},
\end{aligned}
\label{meta2}
\end{equation}
where $m$ and $\eta_{2}$ are the minibatch size of metadata and the step size, respectively. At the same time, the feature means and covariance matrices for all classes are optimized based on the metadata:
\begin{equation}
\begin{aligned}
    {\boldsymbol{\Sigma}}^{(t+1)} & ={\boldsymbol{\Sigma}}^{(t)}-\eta_{2} \frac{1}{m} \sum_{i=1}^{m} \nabla_{\boldsymbol{\Sigma}}\ell_{CE}(\tilde{f}_{\overline{\tilde{\boldsymbol{W}}}(\boldsymbol{\Omega}^{(t)})}(\boldsymbol{x}_{i}^{{meta}}),y_{i}^{{meta}})|_{\boldsymbol{\Sigma}^{(t)}},
\end{aligned}
\label{meta3}
\end{equation}
\begin{equation}
\begin{aligned}
    \boldsymbol{\mu}^{(t+1)}& = \boldsymbol{\mu}^{(t)}-\eta_{2} \frac{1}{m} \sum_{i=1}^{m} \nabla_{\boldsymbol{\mu}}\ell_{CE}(\tilde{f}_{\overline{\tilde{\boldsymbol{W}}}(\boldsymbol{\Omega}^{(t)})}(\boldsymbol{x}_{i}^{{meta}}),y_{i}^{{meta}})|_{\boldsymbol{\mu}^{(t)}}.
\end{aligned}
\label{meta4}
\end{equation}
$\boldsymbol{\Sigma}^{(t)}$ and $\boldsymbol{\mu}^{(t)}$ refer to the covariance matrices and feature means of all classes in step $t$, respectively. Finally, the parameters of the classifier network can be updated with the obtained augmentation strengths $\alpha_{i}^{(t+1)}$:
\begin{equation}
\begin{aligned}
\tilde{\boldsymbol{W}}^{(t+1)}&=\tilde{\boldsymbol{W}}^{(t)}-\eta_{1} \frac{1}{n} \sum_{i=1}^{n}\nabla_{_{\tilde{\boldsymbol{W}}}}\ell_{ICDA}(\tilde{f}(\boldsymbol{x}_{i}),y_{i}; \alpha_{i}^{(t+1)})|_{_{\tilde{\boldsymbol{W}}^{(t)}}}.
\end{aligned}
\label{meta5}
\end{equation}
The steps of Meta-ICDA are presented in Algorithm~1.

\section{Explanation in Regularization View}
\label{secregu}
This section conducts a deeper analysis considering regularization and reveals the ICDA's superiority against three advanced approaches: LA, ISDA, and RISDA. To our knowledge, this is the first time regularization has been used to explain these methods.

\begin{table*}[t]
\centering
\resizebox{1\linewidth}{!}{
\begin{tabular}{c|c|l}
\hline
\textbf{Method}                                                             & \textbf{Regularization term}                                                         &  \makecell[c]{\textbf{Generalization factor}}  \\ \hline
\rowcolor{gray!10} {LA} & ${R_{LA}}{\rm{ = }}\sum_{i = 1}^N {\sum_{c \ne {y_i}} {{{q}_{i,c}}\delta_{c,{i}}} }$ & \begin{tabular}[l]{@{}c@{}}\checkmark Class-wise margin \end{tabular}\\ \hline
{ISDA} & ${R_{ISDA}}{\rm{ = }}\frac{\lambda }{2}\sum_{i = 1}^N {\sum_{c \ne {y_i}} {{{q}_{i,c}}{\Delta \boldsymbol{w}_{c,y_{i}}\boldsymbol{\Sigma}_{y_{i}}\Delta \boldsymbol{w}_{c,y_{i}}^{T}}} }$ & \checkmark Intra-class compactness\\ \hline
\rowcolor{gray!10} {RISDA} & ${R_{RISDA}}{\rm{ = }}\sum_{i = 1}^N {\sum_{c \ne {y_i}} {{{q}_{i,c}}[\alpha \Delta {\boldsymbol{w}_{c,y_{i}}}\sum_{j = 1,j \ne {y_i}}^C {{\varepsilon _{y_i,j}}{\boldsymbol{\mu} _j}}  + \beta \Delta {\boldsymbol{w}_{c,y_{i}}}({\boldsymbol{\Sigma}_{{y_i}}} + \sum_{j = 1,j \ne {y_i}}^C {{\varepsilon _{y_i,j}}} {\boldsymbol{\Sigma} _j})\Delta \boldsymbol{w}_{c,y_{i}}^{T}]} }$ & \begin{tabular}[c]{@{}l@{}} \checkmark Intra-class compactness\\ \checkmark Class-boundary distance \end{tabular}    \\ \hline
{ICDA} & ${R_{ICDA}}{\rm{ = }}\sum\limits_{i = 1}^N \{{\sum\limits_{c \ne {y_i}} {{{q}_{i,c}}[\delta_{c,{i}} + \frac{{\lambda}}{2}{\Delta \boldsymbol{w}_{c,y_{i}}(\boldsymbol{\Sigma}_{y_{i}}\!+\!\sum\limits_{j=1,j\neq y_{i}}^{C}\hat{\alpha}_{i,j}\boldsymbol{\Sigma}_{{j}})\Delta \boldsymbol{w}_{c,y_{i}}^{T}}} }  + {\lambda} \Delta {\boldsymbol{w}_{c,y_{i}}}{\hat{\alpha} _{i,c}}{\boldsymbol{\mu}_c}] - \beta{\alpha_i}{{q}_{i,{y_i}}}\} $ & \begin{tabular}[l]{@{}l@{}} \checkmark Sample-wise/class-wise margin\\ \checkmark Intra-class compactness \\ \checkmark Class-boundary distance\end{tabular}    \\ \hline
\end{tabular}}
\caption{Regularization terms and reflected generalization factors of the four algorithms (LA, ISDA, RISDA, and ICDA).}
\label{table1}
\end{table*}

Using the first-order Taylor expansion of the loss, we have
\begin{equation}
\begin{aligned}
    \ell(\boldsymbol{u}+\Delta \boldsymbol{u}) \approx \ell(\boldsymbol{u})+(\frac{\partial\ell}{\partial\boldsymbol{u}})^{T}\Delta \boldsymbol{u} = \ell(\boldsymbol{u})+(\boldsymbol{q}-\boldsymbol{y})^{T}\Delta \boldsymbol{u},
\end{aligned}
\end{equation}
where $\boldsymbol{q}\!=\!\text{softmax}(\boldsymbol{u})$ and $\boldsymbol{y}$ is the one-hot label. Considering $R\!=\!(\boldsymbol{q}\!-\!\boldsymbol{y})^{T}\Delta \boldsymbol{u}$, the underlying regularizers of all approaches can be derived. The deviation process is presented in Appendix~D. 
The regularizers and the factors affecting the generalization capability are summarized in Table~\ref{table1}.

\begin{figure}[t] 
\centering
\includegraphics[width=1\textwidth]{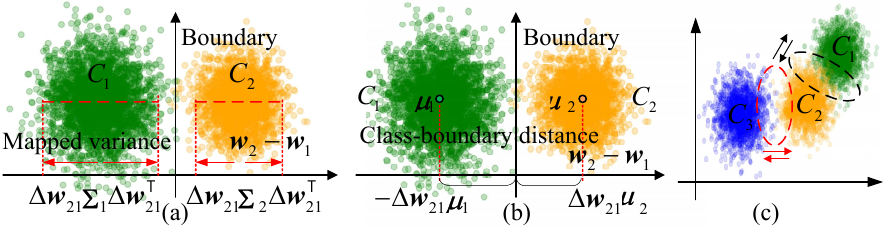}
\vspace{-0.22in}
\caption{Illustrations for the mapped variance (a) and class-boundary distance (b). (c) demonstrates that different samples in the same class should also have distinct augmentation directions.}
\label{fig6}
\end{figure}

Accordingly, $R_{{LA}}$ imposes greater punishment on the predictions ${q}_{i,c}$ ($c\!\neq\!y_i$) with a large $\delta_{c,{i}}$, improving their classification performance. Obviously, tail classes benefit more from LA. Thus, LA is prevalent in handling class imbalance. $R_{{ISDA}}$ contains $\Delta \boldsymbol{w}_{c,y_{i}}\boldsymbol{\Sigma}_{y_{i}}\Delta \boldsymbol{w}_{c,y_{i}}^{T}$, and we prove that it is 
the mapped variance 
from samples of class $y_i$ to the normal vector of the boundary between classes $y_{i}$ and $c$ (Fig.~\ref{fig6}(a)). 
\begin{proof}
If feature $\boldsymbol{h}$ is on the boundary, we have
\begin{equation}
\boldsymbol{w}^{T}_{c}\boldsymbol{h}+{b}_{c} = \boldsymbol{w}_{y_{i}}^{T}\boldsymbol{h}+{b}_{y_{i}}.
\end{equation}
Then, we know that the boundary between classes $c$ and $y_i$ is 
\begin{equation}
\Delta \boldsymbol{w}_{c,y_i}\boldsymbol{h}+\Delta {b}_{c,y_i}=0,
\end{equation}
and $\Delta \boldsymbol{w}_{c,y_{i}}\!=\!\boldsymbol{w}_{c}^{T}\!-\!\boldsymbol{w}_{y_i}^{T}$ refers to the normal direction of the boundary between classes $y_{i}$ and $c$. Thus, the value of the mapping $\Pi(\boldsymbol{h})$ of feature $\boldsymbol{h}$ in class $y$ to $\Delta \boldsymbol{w}_{c,y_{i}}$ is 
\begin{equation}
    \Pi(\boldsymbol{h}) = \Delta \boldsymbol{w}_{c,y_{i}}\boldsymbol{h}+\Delta {b}_{c,y_{i}}.
\end{equation}
The (expected) variance of $\Pi(\boldsymbol{h})$ for $y\!=\!y_i$ denoted by $\Lambda_{y_i}$ is as follows:
\begin{equation}
\begin{aligned}
\Lambda_{c,y_i} &= \mathbbm{E}_{\Pi(\boldsymbol{h}):y=y_{i}}(\Pi(\boldsymbol{h})-\overline{\Pi(\boldsymbol{h})})(\Pi(\boldsymbol{h})-\overline{\Pi(\boldsymbol{h})})^{T}
    \\&=\mathbbm{E}_{\boldsymbol{h}:y=y_i} [\Delta \boldsymbol{w}_{c,y_{i}}(\boldsymbol{h}-\overline{\boldsymbol{h}})(\boldsymbol{h}-\overline{\boldsymbol{h}})^{T}\Delta \boldsymbol{w}^{T}_{c,y_{i}}]
    \\&=\Delta \boldsymbol{w}_{c,y_{i}}\mathbbm{E}_{\boldsymbol{h}:y=y_i}[(\boldsymbol{h}-\overline{\boldsymbol{h}})(\boldsymbol{h}-\overline{\boldsymbol{h}})^{T}]\Delta\boldsymbol{w}^{T}_{c,y_{i}}
    \\&= \Delta \boldsymbol{w}_{c,y_{i}}\boldsymbol{\Sigma}_{y_{i}}\Delta\boldsymbol{w}^{T}_{c,y_{i}},
\end{aligned}
\end{equation}
where $\overline{\Pi(\boldsymbol{h})} = \mathbbm{E}_{\Pi(\boldsymbol{h}):y=y_i}[\Pi(\boldsymbol{h})]$ and $\overline{\boldsymbol{h}} = \mathbbm{E}_{\boldsymbol{h}:y=y_i}[\boldsymbol{h}]$.
\end{proof}

\begin{figure}[t] 
\centering
\includegraphics[width=1\textwidth]{fig/fighebing.pdf}
\vspace{-0.1in}
\caption{(a) illustrates distributions of the margin values trained using the ISDA, RISDA, and ICDA losses. (b) and (c) demonstrate the variation of mapped variances towards normal vectors of boundaries related to (b) and unrelated to (c) the ground-truth class 
on standard CIFAR10 using the ResNet-32 model.}
\label{margin_dis}
\end{figure}

This term will force the model to decrease the mapped variances of class $y_i$ towards the normal vectors of boundaries related to $y_i$, and thus increase intra-class compactness. This explains why ISDA performs well on standard datasets. However, as head classes have large training sizes, the punishment for their ${q}_{i,c}$s and intra-class compactness is large. Xu et al.~\cite{r85} revealed that classes with lower compactness, indicated by large variances, are more challenging and exhibit poorer performance. Consequently, ISDA further impairs the performance of tail classes and enlarges the performance gap between head and tail classes, as the variances of head classes decrease more than those of tail classes,
which is undesirable in long-tailed classification.


The term ${\Delta {\boldsymbol{w}_{c,y_{i}}}({\boldsymbol{\Sigma}_{{y_i}}} + \sum_{j = 1,j \ne {y_i}}^C {{\varepsilon _{y_i,j}}} {{{\boldsymbol{\Sigma} }}_j})\Delta \boldsymbol{w}_{c,y_{i}}^{T}}$ in $R_{RISDA}$ is considered the mapped variances of more classes to the normal vector of the boundary between classes $y_{i}$ and $c$. Therefore, it effectively decreases the intra-class compactnesses of more classes along each boundary and not just the ground-truth class. 
The term $\Delta \boldsymbol{w}_{c,y_{i}}\sum_{j=1,j\neq y_{i}}^{C}\varepsilon_{y_i,j}\boldsymbol{\mu}_{j}$ can actually be divided into two parts: $\varepsilon_{y_i,c}\Delta \boldsymbol{w}_{c,y_{i}}\boldsymbol{\mu}_{c}$ and $\varepsilon_{y_i,c'}\Delta \boldsymbol{w}_{c,y_{i}}\boldsymbol{\mu}_{c^{\prime}}$. Then, we prove that the term $\Delta \boldsymbol{w}_{c,y_{i}}\boldsymbol{\mu}_{c}$ refers to the class-boundary distance between classes $y_i$ and $c$, as illustrated in Fig.~\ref{fig6}(b).
\begin{proof}
The boundary surface between classes $y_i$ and $c$ is
\begin{equation}
    \Delta \boldsymbol{w}_{c,y_{i}}\boldsymbol{h}+\Delta {b}_{c,y_{i}}=0.
\end{equation}
Then, the distance from $\boldsymbol{\mu}_{c}$ to the boundary is
\begin{equation}
    d = \frac{|\Delta \boldsymbol{w}_{c,y_{i}}\boldsymbol{\mu}_{c}+\Delta {b}_{c,y_i}|}{||\Delta\boldsymbol{w}_{c,y_{i}}||}.
\end{equation}
As the feature mean $\boldsymbol{\mu}_{c}$ must be classified correctly, we have $\Delta \boldsymbol{w}_{c,y_{i}}\boldsymbol{\mu}_{c}+\Delta {b}_{c,y_i}>0$. The bias term $\Delta {b}_{c,y_i}$ can be omitted. Thus, we have, when $||\Delta \boldsymbol{w}_{c,y_{i}}||=1$, 
$\Delta \boldsymbol{w}_{c,y_{i}}\boldsymbol{\mu}_{c}$ reflects the distance from $\boldsymbol{\mu}_{c}$ to the boundary between classes $y_{i}$ and $c$. Then, we explain why the term $-\Delta\boldsymbol{w}_{2,1}\boldsymbol{\mu}_1$ in Fig.~\ref{fig6}(b) is negative. 
As the feature mean $\boldsymbol{\mu}_{1}$ must be classified correctly, $\Delta\boldsymbol{w}_{1,2}\boldsymbol{\mu}_{1}>0$ and  $\Delta\boldsymbol{w}_{2,1}\boldsymbol{\mu}_{1}<0$. Therefore, the distance between $\boldsymbol{\mu}_{1}$ and the boundary between classes $C_{1}$ and $C_2$ is the negative of $\Delta\boldsymbol{w}_{2,1}\boldsymbol{\mu}_{1}$.
\end{proof}
The regularization of $\varepsilon_{y_i,c}\Delta \boldsymbol{w}_{c,y_{i}}\boldsymbol{\mu}_{c}$ will then force the boundary to move closer to $\boldsymbol{\mu}_{c}$ and thus increase the class-boundary distance for class $y_i$, benefiting $y_i$. However, regularizing the second part seems unreasonable as $\boldsymbol{\mu}_{c^{\prime}}$ is supposed to have no bias towards both classes $y_{i}$ and $c$. Ideally, this term keeps close to zero rather than having a negative value. Thus, we removed this term from the derived ICDA loss, as stated in Section~\ref{sec32}.


Compared with other methods, 
${R}_{ICDA}$ can force models to simultaneously increase and decrease ${q}_{i,y_{i}}$ and ${q}_{i,c}$, respectively. Thus, sample margins, especially those of hard ones, will be enlarged because the harder the sample, the larger the $\alpha_{i}$. Fig.~\ref{margin_dis}(a) depicts the margin distributions of ISDA, RISDA, and ICDA, demonstrating that ICDA has fewer samples predicted correctly with small margins compared to the other two methods. In addition, the term $\Delta \boldsymbol{w}_{c,y_{i}}\boldsymbol{\mu}_{c}$ in $R_{ICDA}$
increases the class-boundary distance for class $y_i$. Like LA, the term ${q}_{i,c}\delta_{c,{i}}$ further increases the class-wise margins of the tail classes, manifesting that ICDA can deal well with imbalanced classification. Furthermore,
$\Delta \boldsymbol{w}_{c,y_{i}}(\boldsymbol{\Sigma}_{y_{i}}\!+\!\sum_{j=1,j\neq y_{i}}^{C}\hat{\alpha}_{i,j}\boldsymbol{\Sigma}_{{j}})\Delta \boldsymbol{w}_{c,y_{i}}^{T}$is the mapped variance of all relevant classes to the normal vector of the boundary between classes $y_{i}$ and $c$. Since this term is sample-wise, our punishment on the mapped variances is more refined and accurate than the class-wise approaches, enforcing better intra-class compactness. Fig.~\ref{fig6}(c) reveals that although $C_2$ and $C_1$ are more confusing, the samples in the red circle are the most correlated with $C_3$ and cannot be taken seriously by the class-wise approaches. From the results presented in Figs.~\ref{margin_dis}(b) and (c), ICDA decreases the mapped variances not only on the boundaries related to the ground-truth class but also the unrelated ones to a higher degree. 
The $\beta$ and ${\lambda}$ parameters in $R_{ICDA}$ can control the effect of each component.

\section{Experiments}

\begin{table}[t]
\centering
\begin{tabular}{l|cc|cc}
\toprule
Dataset & \multicolumn{2}{c|}{CIFAR10} & \multicolumn{2}{c}{CIFAR100}\\ \hline
Imbalance ratio                        & \multicolumn{1}{c|}{100:1}                          & 10:1    & \multicolumn{1}{c|}{100:1}                          & 10:1                        \\   \hline\hline
Class-balanced CE~\cite{r42}                 & \multicolumn{1}{l|}{72.68\%}                        & 86.90\%    & \multicolumn{1}{l|}{38.77\%}                        & 57.57\%                    \\
Class-balanced Focal~\cite{r42}              & \multicolumn{1}{l|}{74.57\%}                        & 87.48\%   & \multicolumn{1}{l|}{39.60\%}                        & 57.99\%                     \\
LDAM~\cite{r30}                                   & \multicolumn{1}{l|}{73.55\%}                        & 87.32\%   & \multicolumn{1}{l|}{40.60\%}                        & 57.29\%                     \\
LDAM-DRW~\cite{r30}                               & \multicolumn{1}{l|}{78.12\%}                        & 88.37\%   & \multicolumn{1}{l|}{42.89\%}                        & 58.78\%                       \\

LA~\cite{r32}                                      & \multicolumn{1}{l|}{77.67\%}                        & 88.93\%       & \multicolumn{1}{l|}{43.89\%}                        & 58.34\%                  \\
ALA~\cite{r34}                                   & \multicolumn{1}{l|}{77.65\%}                        & 88.32\%     & \multicolumn{1}{l|}{43.67\%}                        & 58.92\%                    \\
De-confound-TDE~\cite{r45}                        & \multicolumn{1}{l|}{\underline{80.60}\%}                        & 88.50\%                 & \multicolumn{1}{l|}{44.10\%}                        & 59.60\%        \\
ISDA~\cite{r8}                         & \multicolumn{1}{l|}{72.55\%}                        & 87.02\%    & \multicolumn{1}{l|}{37.40\%}                        & 55.51\%                   \\
RISDA~\cite{r79}                       & \multicolumn{1}{l|}{{79.89}\%}                        & 89.36\%     & \multicolumn{1}{l|}{\underline{50.16}\%}                        & \underline{62.38}\%                        \\
SGIDA~\cite{wang2024smooth}                       & \multicolumn{1}{l|}{{80.02}\%}                        & 89.25\%     & \multicolumn{1}{l|}{{50.13}\%}                        & {61.90}\%                        \\
{ICDA (Ours)}            & \multicolumn{1}{l|}{{\textbf{81.69}\%}} & {\textbf{90.62}\%}  & \multicolumn{1}{l|}{{\textbf{50.18}\%}} & {\textbf{63.45}\%} \\ \hline
{Meta-Weight-Net~\cite{r46}} & \multicolumn{1}{l|}{{73.57\%}} & {87.55\%}& \multicolumn{1}{l|}{{41.61\%}} & {58.91\%}  \\
{MetaSAug~\cite{r9}}        & \multicolumn{1}{l|}{{{80.54}\%}} & {\underline{89.44}\%} & \multicolumn{1}{l|}{{{46.87}\%}} & {{61.73}\%} \\
LSDA~\cite{pu2024fine}                       & \multicolumn{1}{l|}{\underline{80.67}\%}                        & 89.40\%     & \multicolumn{1}{l|}{\underline{49.35}\%}                        & \underline{62.39}\%                        \\
{Meta-ICDA (Ours)}      & \multicolumn{1}{l|}{{\textbf{82.47}\%}} & {\textbf{91.13}\%} & \multicolumn{1}{l|}{{\textbf{50.96}\%}} & {\textbf{63.97}\%}  \\ 
\bottomrule
\end{tabular}
\caption{Top-1 accuracy on long-tailed CIFAR datasets. Bold and underlined numbers are the best and second-best results.}
\label{table2}
\end{table}

We empirically validate ICDA on several typical learning scenarios \zxl{that require model generalization and robustness} (i.e., biased datasets including both imbalanced and noisy data, subpopulation shifts datasets, generalized long-tailed datasets, and standard datasets) regarding performance and efficiency. Both image and text datasets are evaluated. 
For a fair comparison, Meta-ICDA is only compared when the competitor method utilizes meta-learning. We also visualize the augmented samples in the original input space and the attention of the trained model on several images.
Finally, we conduct ablation studies and sensitivity tests.
Regarding the hyperparameter settings in ICDA, $\lambda^{0}$ is selected in $\{0.1, 0.25, 0.5, 0.75, 1\}$, 
and $\beta$ is set to $0.1$ in all subsections.

\subsection{Experiments on Long-Tailed Datasets}
This section presents experiments on imbalanced datasets, including long-tailed versions of the CIFAR and ImageNet datasets, where models typically exhibit a bias toward head classes.

\subsubsection{Long-Tailed CIFAR Datasets}

\textbf{Settings.} 
Long-tailed CIFAR is the long-tailed version of the CIFAR~\cite{r40} data. 
The original CIFAR10 (CIFAR100) dataset consists of 50,000 images for 10 (100) classes with a balanced class distribution. 
Following Cui et al.~\cite{r42}, we discard some training samples to construct imbalanced
datasets. Two training sets with imbalance ratios of 100:1 and 10:1 are built. 
We train ResNet-32~\cite{r36} with an initial learning rate of $0.1$ and the standard SGD with the momentum of $0.9$ and a weight decay of $5\!\times\!10^{-4}$. The learning rate is decayed by $0.1$ at the 120th and 160th epochs.
As for the meta-learning-based algorithms, the initial learning rate is $0.1$ and it is decayed by $0.01$ at the 160th and 180th epochs following MetaSAug~\cite{r9}. 
We randomly select ten images per class from the training data to construct metadata. 

Several classical and advanced robust losses and data augmentation approaches that are mainly designed for long-tailed classifications are compared, including Class-balanced
CE loss~\cite{r42}, 
Class-balanced Focal loss, LDAM~\cite{r30}, LDAM-DRW~\cite{r30},
ISDA~\cite{r8}, LA~\cite{r32}, ALA~\cite{r34}, RISDA~\cite{r79}, and SGIDA~\cite{wang2024smooth}. Besides, De-confound-TDE~\cite{r45}, which uses causal intervention in training and counterfactual reasoning in inference, is also involved in our comparison. Three meta-learning methods including Meta-Weight-Net~\cite{r46}, MetaSAug~\cite{r9}, and LSDA~\cite{pu2024fine} are also compared.

\begin{figure}[t] 
\centering
\includegraphics[width=0.75\textwidth]{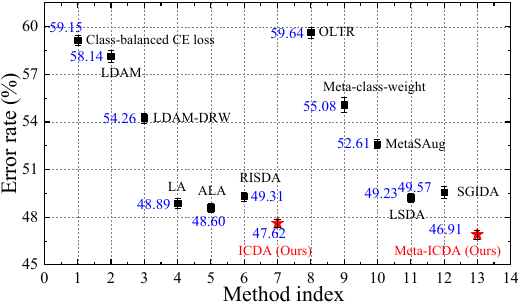}
\vspace{-0.1in}
\caption{Top-1 error rate on ImageNet-LT.}
\label{fig7}
\end{figure}

\begin{table}[bp]
\centering

\begin{tabular}{l|c|cc}
\toprule
{Networks} & {Params} & \multicolumn{2}{c}{Additional cost}    \\ \cline{3-4} 
                          &                         & \multicolumn{1}{l|}{CIFAR10} & CIFAR100 \\ \hline\hline
ResNet-32                 & 0.5M                    & \multicolumn{1}{c|}{6.9\%}   & 6.9\%    \\
ResNet-56                 & 0.9M                    & \multicolumn{1}{c|}{7.1\%}   & 7.2\%    \\
ResNet-110                & 1.7M                    & \multicolumn{1}{c|}{6.8\%}   & 6.7\%    \\\hline
DenseNet-BC-121           & 8M                      & \multicolumn{1}{c|}{5.6\%}   & 5.3\%    \\
DenseNet-BC-265           & 33.3M                   & \multicolumn{1}{c|}{5.3\%}   & 5.1\%    \\\hline
Wide ResNet-16-8          & 11.0M                   & \multicolumn{1}{c|}{6.8\%}     & 7.0\%    \\
Wide ResNet-28-10         & 36.5M                   & \multicolumn{1}{c|}{6.7\%}   & 6.8\%    \\ 
\bottomrule
\end{tabular}
\caption{Additional training time increased by ICDA loss compared with CE loss.}
\label{table3}
\end{table}

\begin{table*}[t]
\centering

\resizebox{1\textwidth}{!}{
\begin{tabular}{l|cc|cc|cc|cc}
\toprule
Dataset &  \multicolumn{2}{c|}{CIFAR10} & \multicolumn{2}{c}{CIFAR100} &  \multicolumn{2}{c|}{CIFAR10} & \multicolumn{2}{c}{CIFAR100} \\\hline
Noise type & \multicolumn{4}{c|}{Flip} & \multicolumn{4}{c}{Uniform} \\ \hline
Noise ratio       & \multicolumn{1}{c|}{20\%}                         & 40\%               & \multicolumn{1}{c|}{20\%}                         & 40\%      & \multicolumn{1}{c|}{40\%}                         & 60\%               & \multicolumn{1}{c|}{40\%}                         & 60\%           \\ \hline\hline
CE loss & \multicolumn{1}{l|}{76.83\%}                        & 70.77\%  & \multicolumn{1}{l|}{50.86\%}                        & 43.01\%        & \multicolumn{1}{l|}{68.07\%}                        & 53.12\%    & \multicolumn{1}{l|}{51.11\%}                        & 30.92\%                 \\
$L_{DMI}$~\cite{zaoshengl}               & \multicolumn{1}{l|}{86.70\%}                        & 84.00\%            & \multicolumn{1}{l|}{62.26\%}                        & 57.23\%       & \multicolumn{1}{l|}{85.90\%}                        & 79.60\%                  & \multicolumn{1}{l|}{63.16\%}                        & 55.37\%                                \\
JoCoR~\cite{zaosheng2}        & \multicolumn{1}{l|}{\underline{90.78}\%}                        & 83.67\%    & \multicolumn{1}{l|}{\underline{65.21}\%}                        & 45.44\%    & \multicolumn{1}{l|}{\underline{89.15}\%}                        & 64.54\%          & \multicolumn{1}{l|}{\underline{65.45}\%}                        & 44.43\%                                  \\
D2L~\cite{r49}               & \multicolumn{1}{l|}{87.66\%}                        & 83.89\%      & \multicolumn{1}{l|}{63.48\%}                        & 51.83\%           & \multicolumn{1}{l|}{85.60\%}                        & 68.02\%                 & \multicolumn{1}{l|}{52.10\%}                        & 41.11\%                        \\
Co-teaching~\cite{r48}       & \multicolumn{1}{l|}{82.83\%}                        & 75.41\%           & \multicolumn{1}{l|}{54.13\%}                        & 44.85\%        & \multicolumn{1}{l|}{74.81\%}                        & 73.06\%              & \multicolumn{1}{l|}{46.20\%}                        & 35.67\%                          \\ 
APL~\cite{r50}               & \multicolumn{1}{l|}{87.23\%}                        & 80.08\%                 & \multicolumn{1}{l|}{59.37\%}                        & 52.98\%            & \multicolumn{1}{l|}{86.49\%}                        & 79.22\%            & \multicolumn{1}{l|}{57.84\%}                        & 49.13\%                         \\
ISDA~\cite{r8}              & \multicolumn{1}{l|}{{88.90}\%}                        & \underline{86.14}\%     & \multicolumn{1}{l|}{{64.36}\%}                        & \underline{59.48}\%   & \multicolumn{1}{l|}{{88.11}\%}                        & \underline{83.12}\%           & \multicolumn{1}{l|}{{65.15}\%}                        & \underline{58.19}\%                         \\
RISDA~\cite{r79}              & \multicolumn{1}{l|}{{85.48}\%}                        & {81.12}\%           & \multicolumn{1}{l|}{{61.81}\%}                        & {54.60}\%        & \multicolumn{1}{l|}{{83.25}\%}                        & {76.31}\%              & \multicolumn{1}{l|}{{54.09}\%}                        & {45.57}\%                              \\
SGIDA~\cite{wang2024smooth}              & \multicolumn{1}{l|}{{84.64}\%}                        & {82.09}\%           & \multicolumn{1}{l|}{{61.56}\%}                        & {55.23}\%        & \multicolumn{1}{l|}{{84.47}\%}                        & {75.89}\%              & \multicolumn{1}{l|}{{56.13}\%}                        & {46.11}\%                              \\


ICDA (Ours)              & \multicolumn{1}{l|}{{\textbf{91.81}\%}} & {\textbf{88.76}\%}   & \multicolumn{1}{l|}{{\textbf{66.85}\%}} & {\textbf{61.57}\%}  & \multicolumn{1}{l|}{{\textbf{90.23}\%}} & {\textbf{84.91}\%}         & \multicolumn{1}{l|}{{\textbf{67.24}\%}} & {\textbf{60.26}\%} \\ \hline
MentorNet~\cite{r51}         & \multicolumn{1}{l|}{86.36\%}                        & 81.76\%              & \multicolumn{1}{l|}{61.97\%}                        & 52.66\%        & \multicolumn{1}{l|}{87.33\%}                        & 82.80\%                 & \multicolumn{1}{l|}{61.39\%}                        & 36.87\%                          \\
Meta-Weight-Net~\cite{r46}   & \multicolumn{1}{l|}{90.33\%}                        & 87.54\%              & \multicolumn{1}{l|}{64.22\%}                        & 58.64\%            & \multicolumn{1}{l|}{89.27\%}                        & 84.07\%            & \multicolumn{1}{l|}{\underline{67.73}\%}                        & 58.75\%                 \\

MetaSAug~\cite{r9}          & \multicolumn{1}{l|}{\underline{90.42}\%}                        & 87.73\%   & \multicolumn{1}{l|}{\underline{66.47}\%}                        & {61.43}\%      & \multicolumn{1}{l|}{\underline{89.32}\%}                        & \underline{84.65}\%       & \multicolumn{1}{l|}{{66.50\%}} & {\underline{59.84}\%}                   \\
LSDA~\cite{pu2024fine}          & \multicolumn{1}{l|}{{90.36}\%}                        & 87.85\%   & \multicolumn{1}{l|}{{66.04}\%}                        & {62.11}\%      & \multicolumn{1}{l|}{{89.01}\%}                        & {84.20}\%       & \multicolumn{1}{l|}{{66.63\%}} & {{59.67}\%}                   \\
Meta-ICDA (Ours)       & \multicolumn{1}{l|}{{\textbf{92.46}\%}} & {\textbf{90.21}\%} & \multicolumn{1}{l|}{{\textbf{67.54}\%}} & {\textbf{63.26}\%} & \multicolumn{1}{l|}{{\textbf{91.14}\%}} & {\textbf{85.86}\%}   & \multicolumn{1}{l|}{{\textbf{68.92}\%}} & {\textbf{61.80}\%}\\ 
\bottomrule
\end{tabular}}
\caption{Top-1 accuracy on CIFAR datasets with uniform and flip noises.}
\label{table4}
\end{table*}

\textbf{Results.}
Table~\ref{table2} reports the results on long-tailed CIFAR data, which are divided into two groups according to the usage of meta-learning. 
The results reveal that ICDA 
significantly outperforms other reweighting, solely logit adjustment, and implicit semantic augmentation methods, demonstrating that our sample-wise counterfactual augmentation strategy deals well with long-tailed classification. Although ICDA and RISDA achieve comparable performance on CIFAR100 with an imbalance ratio of 100:1, ICDA outperforms RISDA in other cases.
Additionally, ICDA consistently surpasses De-confound-TDE, which uses causal intervention in training and counterfactual reasoning in inference. Our Meta-ICDA achieves state-of-the-art performance compared to all approaches.  

To evaluate efficiency, we record the additional training time for ICDA and compare it against CE and ISDA. Table~\ref{table3} reports the additional training time introduced by ICDA loss compared with CE loss on various backbones. The additional time introduced by ISDA loss compared with CE loss can be seen in the ISDA paper~\cite{r8}. The results reveal that only a little time is increased by ICDA loss,
and the values of training time for ICDA and ISDA are nearly equivalent.

\subsubsection{Long-Tailed ImageNet Dataset}

\textbf{Settings.}
ImageNet~\cite{r41} is a benchmark visual recognition dataset, which contains 1,281,167 training images and
50,000 validation images. 
Liu et al.~\cite{r97} built the long-tailed version of ImageNet, which is denoted as ImageNet-LT. After
discarding some training samples, ImageNet-LT remains
115,846 training examples in 1,000 classes. The imbalance
ratio of ImageNet-LT is 256:1. Following MetaSAug~\cite{r9}, we adopt the original validation set to test
methods. Ten images per class, which are selected from the balanced validation set compiled by Liu et al.~\cite{r97} are utilized to construct our metadata. ResNet-50~\cite{r36} is used as the backbone network. The learning rate is decayed by $0.1$ at the 60th and 80th epochs. The batch size is set to $64$. Only the last fully connected layer is finetuned for efficiency. Methods designed for long-tailed classification including Class-balanced CE loss~\cite{r42}, OLTR~\cite{r97}, LDAM~\cite{r30}, LDAM-DRW~\cite{r30}, LA~\cite{r32}, ALA~\cite{r34}, RISDA~\cite{r79}, Meta-class-weight~\cite{r46}, MetaSAug~\cite{r9}, SGIDA~\cite{wang2024smooth}, and LSDA~\cite{pu2024fine} are compared. 

\textbf{Results.}
Fig.~\ref{fig7} highlights that ICDA achieves good performance among the robust losses. Meta-ICDA significantly outperforms all competitor methods, including the meta semantic augmentation approach, proving that our proposed approach is more effective on long-tailed data. 

\subsection{Experiments on Noisy Datasets}

\textbf{Settings.} 
Following Shu et al~\cite{r46}, two settings of corrupted labels are adopted, namely, uniform and pair-flip noise
labels; 
1,000 images with clean labels in the
validation set are selected as the metadata. Wide ResNet-28-10 (WRN-28-10)~\cite{r38} and ResNet-32~\cite{r36} are adopted as
the classifiers for the uniform and pair-flip noises, respectively. The initial learning rate and batch size are set to $0.1$ and $128$, respectively. For ResNet, standard SGD with the momentum of $0.9$ and a weight decay of $1\!\times\!10^{-4}$ is utilized. For Wide ResNet, standard SGD with the momentum of $0.9$ and a weight decay of $5\!\times\!10^{-4}$ is utilized. For the meta-learning-based algorithms, the initial learning rate is set to $0.1$ and is subsequently decayed by a factor of $0.01$ at the 160th and 180th epochs, following the protocol established in MetaSAug~\cite{r9}.

\begin{table*}[t]
\centering

\resizebox{1\textwidth}{!}{
\begin{tabular}{l|cc|cc|cc|cc}
\toprule
         Dataset & \multicolumn{2}{c|}{CelebA} & \multicolumn{2}{c}{CMNIST} & \multicolumn{2}{c|}{Waterbirds} & \multicolumn{2}{c}{CivilComments}\\ \hline\hline
   Method       & Avg.         & Worst        & Avg.          & Worst  & Avg. & Worst & Avg. & Worst      \\ \hline
UW~\cite{r57}        & 92.9\%       & 83.3\%       & 72.2\%       & 66.0\%  &  {95.1}\%          & 88.0\%         & 89.8\%            & 69.2\%               \\
IRM~\cite{r58}       & 94.0\%       & 77.8\%       & 72.1\%       & 70.3\%  & 87.5\%          & 75.6\%         & 88.8\%            & 66.3\%             \\
IB-IRM~\cite{r59}    & 93.6\%       & 85.0\%       & 72.2\%       & 70.7\% & 88.5\%           & 76.5\%         & 89.1\%            & 65.3\%                \\
V-REx~\cite{r60}     & 92.2\%       & 86.7\%       & 71.7\%       & 70.2\%  & 88.0\%           & 73.6\%         & 90.2\%            & 64.9\%                \\
CORAL~\cite{DORAL}     & 93.8\%       & 76.9\%       & 71.8\%       & 69.5\%  & 90.3\%           & 79.8\%         & 88.7\%            & 65.6\%                \\
GroupDRO~\cite{r54}  & 92.1\%       & 87.2\%       & 72.3\%       & 68.6\%  & 91.8\%           & \underline{90.6}\%         & 89.9\%            & 70.0\%                  \\
DomainMix~\cite{r65} & 93.4\%       & 65.6\%       & 51.4\%       & 48.0\%  & 76.4\%           & 53.0\%         & 90.9\%            & 63.6\%               \\
Fish~\cite{r66}      & 93.1\%       & 61.2\%       & 46.9\%       & 35.6\%  & 85.6\%           & 64.0\%         & 89.8\%            & 71.1\%                 \\
LISA~\cite{r57}      & 92.4\%       & \underline{89.3}\%       & {74.0}\%       & \underline{73.3}\%      & 91.8\%           & 89.2\%         & 89.2\%            & \underline{72.6}\%         \\
ICDA (Ours)      & 93.3\%       & \textbf{90.7}\%       & {76.1}\%       & \textbf{75.3}\%    & 92.9\%           & \textbf{90.7}\%         & {91.1}\%            & \textbf{73.5}\%       \\ 
\bottomrule
\end{tabular}}
\caption{Average and worst-group accuracy on subpopulation shifts datasets.}
\label{table6}
\end{table*}

\begin{table*}[t]
\centering

\resizebox{1\textwidth}{!}{
\begin{tabular}{l|cc|cc|cc}
\toprule
Protocol     & \multicolumn{2}{c|}{CLT}  & \multicolumn{2}{c|}{GLT} & \multicolumn{2}{c}{ALT} \\ \hline\hline
Method   & Acc.        & Prec.      & Acc.        & Prec.      & Acc.        & Prec.      \\ \hline
CE loss    &42.52\%  &47.92\%   & 34.75\%       & 40.65\%      & 41.73\%       & 41.74\%      \\
cRT~\cite{r68}       &45.92\% &45.34\%     & 37.57\%       & 37.51\%     & 41.59\%       & 41.43\%      \\
LWS~\cite{r68}        &46.43\% &45.90\%    & 37.94\%       & 38.01\%      & 41.70\%       & 41.71\%      \\
De-confound-TDE~\cite{r45} & 45.70\% &44.48\% &37.56\%       & 37.00\%      & 41.40\%       & 42.36\%      \\
LA~\cite{r32}&   46.53\% & 45.56\%  & 37.80\%       & 37.56\%      & 41.73\%          & 41.74\%         \\
BBN~\cite{r69}     &  46.46\% & 49.86\%     & 37.91\%       & 41.77\%      & 43.26\%       & 43.86\%      \\
LDAM~\cite{r30}  &      {46.74}\% & 46.86\%   & {38.54}\%       & 39.08\%      & 42.66\%       & 41.80\%      \\
IFL~\cite{r67}        & 45.97\% & 52.06\%    & 37.96\%       & 44.47\%      & 45.89\%       & {46.42}\%      \\
RandAug~\cite{r71}     &46.40\% & \underline{52.13}\%      & 38.24\% & \underline{44.74}\%   & {46.29}\% &46.32\%   \\
ISDA~\cite{r8}     & {47.66}\% & 51.98\%& \underline{39.44}\%   &  44.26\%  & \underline{47.62}\% & \underline{47.46}\% \\
RISDA~\cite{r79}     & {49.31}\% & 50.64\%& {38.45}\%   &  42.77\%  & {47.33}\% & {46.33}\% \\
SGIDA~\cite{wang2024smooth}     & \underline{49.53}\% & 51.56\%& {38.76}\%   &  43.21\%  & {47.54}\% & {46.58}\% \\
ICDA (Ours)      & \textbf{52.11}\% &     \textbf{55.05}\% &  \textbf{42.73}\%         &  \textbf{47.49}\%           &    \textbf{50.52}\%          &   \textbf{49.68}\%         \\ \hline
MetaSAug~\cite{r9}     & 50.53\% &     \underline{55.21}\% &  {41.27}\%         &   \underline{47.38}\%      &    {49.12}\%         &   \underline{48.56}\%          \\ 
LSDA~\cite{pu2024fine}     & \underline{50.78}\% &     55.09\% &  \underline{41.35}\%         &   {47.26}\%      &    \underline{49.20}\%         &   {48.44}\%          \\ 
Meta-ICDA (Ours)      & \textbf{52.76}\% &     \textbf{56.71}\% &  \textbf{44.15}\%         &   \textbf{49.32}\%      &    \textbf{51.74}\%         &   \textbf{51.43}\%         \\ 
\bottomrule
\end{tabular}}
\caption{Top-1 accuracy and precision of the CLT, GLT, and ALT protocols on the ImageNet-GLT benchmark.}

\label{table7}
\end{table*}

Several robust loss functions, including 
Information-theoretic Loss ($L_{DMI}$)~\cite{zaoshengl}, JoCoR~\cite{zaosheng2}, Co-teaching~\cite{r48}, D2L~\cite{r49}, and APL~\cite{r50} are compared. The meta-learning-based methods, including MentorNet~\cite{r51}, Meta-Weight-Net~\cite{r46}, and LSDA~\cite{pu2024fine}, are also involved in comparison. We also compared our proposed ICDA with three implicit data augmentation methods, including ISDA~\cite{r8}, RISDA~\cite{r79}, MetaSAug~\cite{r9}, and SGIDA~\cite{wang2024smooth}. 


\textbf{Results.} Table~\ref{table4} reports the results of CIFAR data with
flip and uniform noise, respectively. 
ICDA notably surpasses all competitor approaches, including robust loss functions and the class-level implicit data augmentation approaches. Besides, Meta-ICDA achieves state-of-the-art performance compared with other meta-learning-based manners, manifesting that our proposed method can effectively improve the generalization and robustness of models on noisy data.

\begin{table*}[t]
\centering

\resizebox{1\textwidth}{!}{
\begin{tabular}{l|cc|cc|cc}
\toprule
Protocol     & \multicolumn{2}{c|}{CLT}  & \multicolumn{2}{c|}{GLT} & \multicolumn{2}{c}{ALT} \\ \hline\hline
Method   & Acc.        & Prec.      & Acc.        & Prec.      & Acc.        & Prec.      \\ \hline
CE loss    &72.34\% & 76.61\% & 63.79\% & 70.52\% & 50.17\% & 50.94\%      \\
cRT~\cite{r68}       &73.64\% & 75.84\% &  64.69\% & 68.33\% & 49.97\% & 50.37\%      \\
LWS~\cite{r68}        &72.60\% & 75.66\% & 63.60\% & 68.81\% & 50.14\% & 50.61\%      \\
De-confound-TDE~\cite{r45} & 73.79\% & 74.90\% & 66.07\% & 68.20\% &  50.76\% & 51.68\%      \\
LA~\cite{r32}&   75.50\% & 76.88\% & 66.17\% & 68.35\% & 50.17\% & 50.94\%         \\
BBN~\cite{r69}      &  73.69\% & 77.35\% & 64.48\% & 70.20\% & 51.83\% & 51.77\%     \\
LDAM~\cite{r30}  &      75.57\% &  77.70\% & 67.26\% & 70.70\% & \underline{55.52}\% & \underline{56.21}\%      \\
IFL~\cite{r67}        & 74.31\% & 78.90\% & 65.31\% & 72.24\% & 52.86\% & 53.49\%      \\
RandAug~\cite{r71}     &{76.81}\% & \underline{79.88}\% & \underline{67.71}\% & {72.73}\% & 53.69\% & {54.71}\%   \\
ISDA~\cite{r8}   &\underline{77.32}\%  & 79.23\% & 67.57\% & {72.89}\% & {54.43}\% &  54.62\%  \\
RISDA~\cite{r79}   &{76.34}\%  & 79.27\% & 66.85\% & {72.66}\% & {54.58}\% &  53.98\%  \\
SGIDA~\cite{wang2024smooth}   &{76.87}\%  & 79.43\% & 67.06\% & \underline{72.90}\% & {55.14}\% &  54.27\%  \\
ICDA (Ours)      & \textbf{78.82}\% & \textbf{81.33}\%    &     \textbf{68.78}\%       &   \textbf{74.29}\%         &   \textbf{56.48}\% &       \textbf{57.81}\%     \\ 
\bottomrule
\end{tabular}}
\caption{Top-1 accuracy and precision of the CLT, GLT, and ALT protocols on the MSCOCO-GLT benchmark.}
\label{table8}
\end{table*}

\subsection{Experiments on Subpopulation Shifts Datasets}
\textbf{Settings.} Four subpopulation shifts datasets are evaluated, including CMNIST, Waterbirds~\cite{r54}, CelebA~\cite{r55}, and CivilComments~\cite{r56}, in which the domain information is highly spuriously correlated with the labels. Detailed descriptions of the datasets are shown in Appendix~E. In the subsequent trials, 
ResNet-50
is utilized as the backbone network for the first three image datasets, while DistilBert~\cite{r39} is adopted for the text set CivilComments.  The initial learning rates for CMNIST and Waterbirds are $1\times 10^{-3}$, while those for CelebA and Civilcomments are $1\times 10^{-4}$ and $1\times 10^{-5}$, respectively. The values of weight decay are $1\times 10^{-4}$ for CMNIST, Waterbirds, and CelebA, and $0$ for CivilComments. The values of batch size for CMNIST, Waterbirds, and CelebA are $16$, and that for Civilcomments is $8$. For the three image classification datasets, the SGD optimizer is utilized, while Adam is utilized for CivilComments.

Robust learning methods, including IRM~\cite{r58}, IB-IRM~\cite{r59}, V-REx~\cite{r60},  CORAL~\cite{DORAL}, GroupDRO~\cite{r54}, DomainMix~\cite{r65}, Fish~\cite{r66}, and LISA~\cite{r57}, are involved into comparison. Upweighting (UW) is suitable for subpopulation shifts, so we also use it for comparison.
We only compare ICDA with other methods for fair comparisons, as all these approaches do not rely on meta-learning. Following Yao et al.~\cite{r57}, the worst-group accuracy is used to compare the performance of all compared baselines.

\textbf{Results.}
Table~\ref{table6} reports the results of the four subpopulation shift datasets. The performance of methods that learn invariant
predictors with explicit regularizers, e.g., IRM, IB-IRM, and V-REx, is not consistent across datasets. For example, V-REx outperforms IRM on CelebA, but it fails to
achieve better performance than IRM on CMNIST, Waterbirds, and CivilComments. Opposing, ICDA 
consistently achieves an appealing performance on all datasets,
demonstrating ICDA's effectiveness in breaking spurious correlations and achieving invariant feature learning. Although ICDA and GroupDRO achieve similar performance on Waterbirds, ICDA far exceeds GroupDRO on the other three datasets.

\subsection{Experiments on Generalized Long-Tailed Datasets}
\textbf{Settings.} 
Tang et al.~\cite{r67} proposed a novel learning problem, namely, generalized long-tailed classification, in which two new benchmarks, including MSCOCO-GLT and ImageNet-GLT, were proposed. Each benchmark has three protocols, i.e., CLT, ALT, and GLT, in which class distribution, attribute distribution, and both class and attribute distributions are changed from training to testing, respectively. More details of the two benchmarks can be seen in~\cite{r67}. 
The training and testing configurations follow those in the IFL~\cite{r67} paper. ResNeXt-50~\cite{r37} is used as the backbone network for all methods except for BBN~\cite{r69}. Both Top-1 accuracy and precision are presented. All models are trained with a batch size of $256$ and an initial learning rate of $0.1$. SGD optimizer is utilized with a weight decay of $5\times 10^{-4}$ and the momentum of $0.9$. 
Here, Meta-ICDA is exclusively evaluated on the ImageNet-GLT benchmark. 
To collect the attribute-wise balanced metadata, images from each class in a balanced validation set compiled by Liu et al.~\cite{r97} are clustered into 6 groups by KMeans using a pre-trained
ResNet-50 model.
From each group and class, 10 images are sampled to construct the metadata.

As for the compared methods, we studied the two-stage re-sampling
methods, including cRT~\cite{r68} and LWS~\cite{r68},
posthoc distribution adjustment methods including De-confound-TDE~\cite{r45} and LA~\cite{r32}, multi-branch models with diverse sampling
strategies like BBN~\cite{r69}, invariant feature learning methods like IFL~\cite{r67}, and
reweighting loss functions like LDAM~\cite{r30}.
We also compare some data augmentation methods, including RandAug~\cite{r71}, ISDA~\cite{r8}, RISDA~\cite{r79}, MetaSAug~\cite{r9}, SGIDA~\cite{wang2024smooth}, and LSDA~\cite{pu2024fine}.

\begin{table}[t]
\centering

\begin{tabular}{l|c|c|c|c}
\toprule
Backbone &  \multicolumn{2}{c|}{ResNet-110} & \multicolumn{2}{c}{WRN-28-10} \\ \hline
Dataset       & CIFAR10 & CIFAR100 & CIFAR10 & CIFAR100  \\ \hline\hline
Large Margin~\cite{r89}   & 6.46\%   & 28.00\%   & 3.69\%   & 18.48\%   \\
Disturb Label~\cite{r90}  & 6.61\%   & 28.46\%   & 3.91\%   & 18.56\%   \\
Focal loss~\cite{r43}     & 6.68\%   & 28.28\%   & 3.62\%   & 18.22\%  \\
Center loss~\cite{r91}    & 6.38\%   & 27.85\%    & 3.76\%   & 18.50\%   \\
Lq loss~\cite{r92}        & 6.69\%   & 28.78\%   & 3.78\%   & 18.43\%  \\\hline
WGAN~\cite{r93}           & 6.63\%   & -         & 3.81\%   & -   \\
CGAN~\cite{r94}           & 6.56\%   & 28.25\%  & 3.84\%   & 18.79\%    \\
ACGAN~\cite{r95}          & 6.32\%   & 28.48\%  & 3.81\%   & 18.54\%   \\
infoGAN~\cite{r96}        & 6.59\%   & 27.64\%    & 3.81\%   & 18.44\% \\\hline
ISDA~\cite{r8} & \underline{5.98}\%   & \underline{26.35}\% & \underline{3.58}\%   & \underline{17.98}\%   \\
RISDA~\cite{r79} & {6.47}\%   & {28.42}\% & {3.79}\%   & {18.46}\%    \\
SGIDA~\cite{wang2024smooth} & {6.01}\%   & {27.85}\% & {3.66}\%   & {18.65}\%    \\
ICDA (Ours)           & \textbf{4.89}\%   & \textbf{25.21}\%    & \textbf{3.01}\%   & \textbf{17.03}\%\\ 
\bottomrule
\end{tabular}
\caption{Top-1 error rate on standard CIFAR datasets.}

\label{table9}
\end{table}

\begin{figure*}[bp] 
\centering
\includegraphics[width=1\textwidth]{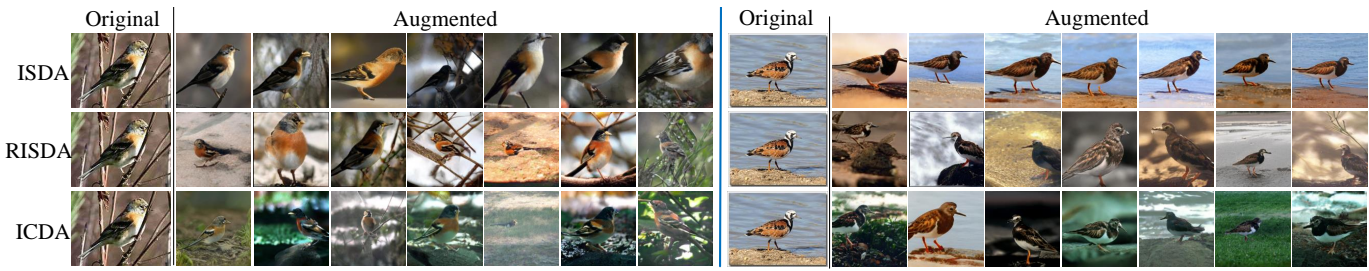}
\vspace{-0.2in}
\caption{Visualization of images augmented by ISDA, RISDA, and ICDA.
}
\label{fig8}
\end{figure*}


\begin{figure*}[t] 
\centering
\includegraphics[width=1\textwidth]{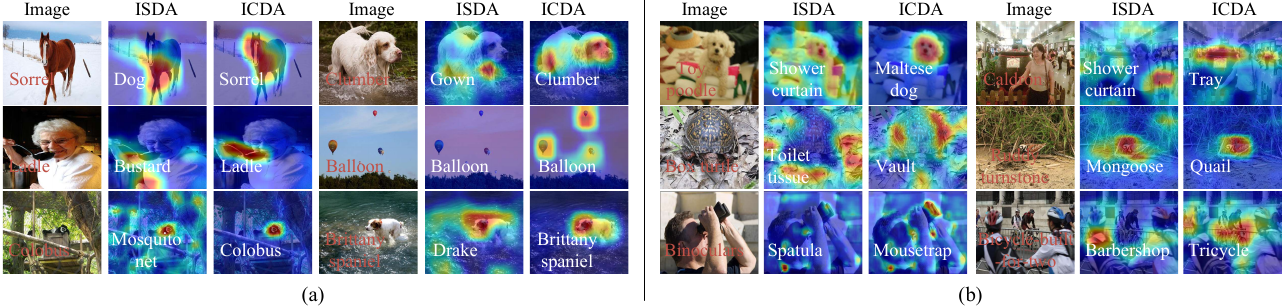}
\vspace{-0.2in}
\caption{(a) Visualization of the regions that the model uses for making predictions. Blue and red imply that the region is indecisive and very discriminative, respectively. White texts are the predicted labels. Red texts are the ground-truth labels of the images. (b) More results of model visualization, in which both ISDA and ICDA make false predictions. 
}
\label{fig9}
\vspace{-0.1in}
\end{figure*}


\textbf{Results.} 
Tables~\ref{table7} and~\ref{table8} report the results of the three protocols for ImageNet-GLT and MSCOCO-GLT, respectively, some of which are from the IFL~\cite{r67} paper. 
ICDA notably improves model performance in all three protocols, demonstrating that it can well break the spurious associations caused by imbalanced attribute and class distributions, while the
majority of previous LT algorithms using rebalancing strategies fail
to improve the robustness against the attribute-wise bias. Additionally, we found that augmentation methods generally perform better than other
long-tailed transfer learning approaches on GLT protocols.

\subsection{Experiments on Standard CIFAR Datasets}

\textbf{Settings.} 
To verify that ICDA has a good augmentation effect, it is compared with a number of advanced methods ranging from robust loss functions (i.e., Large Margin~\cite{r89}, Dsitrub Label~\cite{r90}, Focal loss~\cite{r43}, Center loss~\cite{r91}, and Lq loss~\cite{r92}) to explicit (i.e., WGAN~\cite{r93}, CGAN~\cite{r94}, ACGAN~\cite{r95}, and infoGAN~\cite{r96}) and implicit (i.e., ISDA~\cite{r8}, RISDA~\cite{r79}, and SGIDA~\cite{wang2024smooth}) augmentation methods on standard CIFAR data. ResNet-110 and WRN-28-10 models are utilized. Regarding the hyperparameter settings, the initial learning rate and the batch size are set to $0.1$ and $128$, respectively. For ResNet, standard SGD with the momentum of $0.9$ and a weight decay of $1\!\times\!10^{-4}$ is utilized. For Wide ResNet, standard SGD with the momentum of $0.9$ and a weight decay of $5\!\times\!10^{-4}$ is utilized. 
The learning rate is decayed by $0.1$ at the 120th and 160th epochs.

\textbf{Results.} 
The results are reported in Table~\ref{table9}.
ICDA achieves the best performance compared with other explicit and implicit augmentation approaches. Moreover, GAN-based methods perform poorly on CIFAR100 due to a limited training size. Additionally, these methods impose excessive calculations and decrease training efficiency. 
Although ISDA affords lower error and is more efficient than GAN-based schemes, it can not surpass ICDA, as ICDA assists the models in breaking spurious correlations of models.

\subsection{Visualization Results}

Following ISDA's visualization manner, we map the augmented features back into the pixel space.
The corresponding results are presented in Fig.~\ref{fig8}, highlighting that ICDA can generate more diverse and meaningful counterfactual images and notably alter the non-intrinsic attributes, e.g., scene contexts and viewpoints, compared with ISDA and RISDA. 

\begin{figure}[t] 
\centering
\includegraphics[width=1\textwidth]{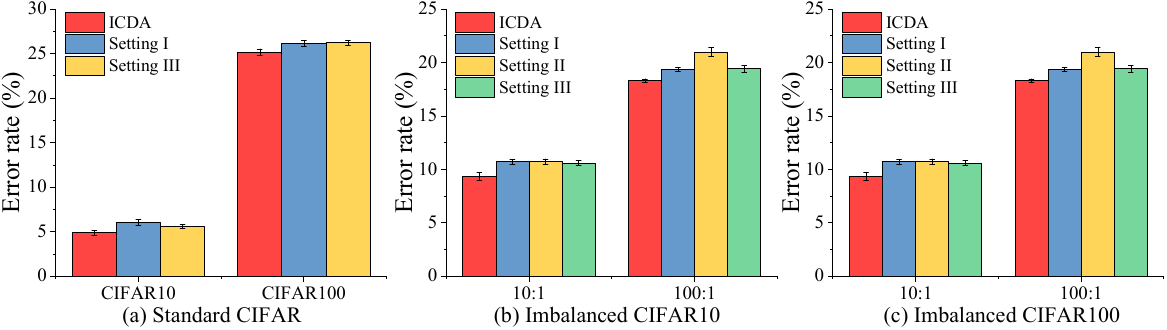}
\vspace{-0.2in}
\caption{Results of ablation studies on standard and imbalanced CIFAR data.}
\label{fig11}
\end{figure}

\begin{figure}[t] 
\centering
\includegraphics[width=0.85\textwidth]{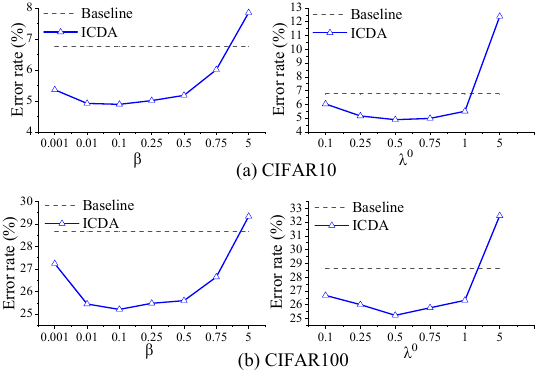}
\vspace{-0.1in}
\caption{Results of the sensitivity tests on standard CIFAR datasets. 
}
\label{fig10}
\end{figure}

Additionally, Grad-CAM~\cite{r72} is utilized to visualize the regions that models use for making predictions.
Fig.~\ref{fig9}(a) manifests that ISDA focuses on the background or other nuisances for false predictions, while ICDA focuses tightly on the causal regions corresponding to the object, assisting models in making correct classifications.  
For example, for the image of ``Brittany spaniel", ISDA utilizes spurious context ``Water", so its prediction is ``Drake", while the model trained with ICDA attends more to the dog, contributing to a correct prediction. Therefore, in addition to performance gains, the ICDA predictions are made for the right reasons. Fig.~\ref{fig9}(b) presents some images that are wrongly classified by both ISDA and ICDA. Although both methods produce incorrect predictions, ICDA enables the model to focus more effectively on causal attributes by disrupting the spurious associations between nuisance factors and class labels. Additional visualization results are provided in Appendix~F.

\subsection{Ablation and Sensitivity Studies}
To get a better understanding of the effect of varying components, we evaluate the following three settings of ICDA. Setting~I: Without the covariance matrices and feature means of the other classes, i.e., $\alpha_{i,c}=0$. Setting~II: Without the class-level logit perturbation term, i.e., removing $\delta_{c, i}$. Setting~III: Without the sample-level logit perturbation term, i.e., removing $\beta\alpha_{i}$. Since the proportion of each class is the same on standard data, we only evaluate Settings~I and~III on the standard data. The ablation results are presented in Fig.~\ref{fig11}, revealing that all three components are crucial and necessary for imbalanced data. Additionally, the statistical information of the other classes and the sample-level perturbation term are critical for standard data. Without each of them, the performance of ICDA will be weakened. 

To study how the hyperparameters in ICDA (i.e., $\lambda^{0}$ and $\beta$) affect our method's performance, several sensitivity tests are conducted, where ResNet-110 is used as the backbone network. The corresponding results on standard CIFAR data are shown in Fig.~\ref{fig10}, revealing that ICDA achieves
superior performance for $0.01\le\beta\le 0.5$ and $0.25\le\lambda^{0}\le 1$. When $\beta$ and $\lambda^{0}$ are too large, the model is easier to overfit and underfit, respectively. Empirically, we recommend $\beta=0.1$ and $\lambda^{0} = 0.5$ for a naive implementation or a starting point of hyperparameter searching.


\section{Conclusion}
This study proposes a novel sample-wise implicit counterfactual data augmentation (ICDA) method aimed at mitigating spurious correlations and enhancing the stability of model predictions.
Our method can be formulated as a novel robust loss, easily adopted by any classifier, and is considerably more efficient than explicit augmentation approaches. Two manners, including direct quantification and meta-learning, are introduced to learn the key parameters in the robust loss. 
Furthermore, the regularization analysis demonstrates that ICDA improves intra-class compactness, class and sample-wise margins, and class-boundary distances.
Extensive experimental comparison and visualization results on several typical learning scenarios demonstrate the proposed method's effectiveness and efficiency.

\biboptions{numbers,sort&compress}
\bibliographystyle{elsarticle-num}
\bibliography{ijcai2023}

\end{document}